\documentclass{article}

\usepackage{microtype}
\usepackage{graphicx}
\usepackage{subfigure}
\usepackage{booktabs} 

\usepackage{multirow}
\usepackage{url}            
\usepackage{bbm}
\usepackage{graphicx}
\graphicspath{{./figures/}}
\usepackage{amsmath}
\usepackage{color}

\usepackage{hyperref}

\usepackage[accepted]{icml2019}

\icmltitlerunning{Single-Path NAS: Device-Aware Efficient ConvNet Design}

\begin{document}

\twocolumn[
\icmltitle{Single-Path NAS: Device-Aware Efficient ConvNet Design}



\icmlsetsymbol{equal}{*}

\begin{icmlauthorlist}
\icmlauthor{Dimitrios Stamoulis}{cmu,equal}
\icmlauthor{Ruizhou Ding}{cmu}
\icmlauthor{Di Wang}{msr}
\icmlauthor{Dimitrios Lymberopoulos}{msr}
\icmlauthor{Bodhi Priyantha}{msr}
\icmlauthor{Jie Liu}{hit}
\icmlauthor{Diana Marculescu}{cmu}
\end{icmlauthorlist}

\icmlaffiliation{cmu}{Department of ECE, Carnegie Mellon University, Pittsburgh, PA, USA}
\icmlaffiliation{msr}{Microsoft, Redmond, WA, USA}
\icmlaffiliation{hit}{Harbin Institute of Technology, Harbin, China}

\icmlcorrespondingauthor{Dimitrios Stamoulis}{dstamoul@andrew.cmu.edu}

\icmlkeywords{Neural Architecture Search, Hardware-aware ConvNets}

\vskip 0.3in
]



\printAffiliationsAndNotice{*Extended abstract of ODML-CDNNR 2019 presentation (required 
non-archival arxiv.org version). Full paper can be found 
in~\cite{stamoulis2019single}.} 

\begin{abstract}
Can we automatically design a Convolutional Network (ConvNet) with the
highest image classification accuracy under the latency constraint 
of a mobile device? Neural Architecture Search (NAS) for ConvNet design
is a challenging problem due to the 
combinatorially large design space and search time (at least 200 GPU-hours). 
To alleviate this complexity, we propose \textit{Single-Path NAS}, a novel 
differentiable NAS method for designing device-efficient ConvNets 
in \textbf{less than 4 hours}. 1.~\textbf{Novel NAS formulation}:
our method introduces a \textbf{single-path}, over-parameterized ConvNet to 
encode all architectural decisions with shared 
convolutional kernel parameters. 
2.~\textbf{NAS efficiency}: Our method decreases the NAS search cost 
down to \textbf{8 epochs} (30 TPU-hours), \textit{i.e.}, up to
\textbf{5,000$\times$ faster} compared to prior work.  
3.~\textbf{On-device image classification}: \textit{Single-Path NAS} achieves 
$74.96\%$ top-1 accuracy on 
ImageNet with 79ms inference latency on a Pixel 1 phone, which is 
state-of-the-art accuracy compared to NAS methods 
with similar latency ($\leq 80ms$).
\end{abstract}

\section{Introduction}

``\textit{Is it possible to reduce the NAS search cost 
down to only \textit{few hours}?}'' NAS methods have revolutionized the design of 
ConvNets~\cite{zoph2017learning},
yielding state-of-the-art results in deep learning applications~\cite{real2018regularized}.
NAS has a profound impact on the design of hardware-efficient 
ConvNets for on-device computer vision, \textit{e.g.}, under 
inference latency constraints on a mobile device~\cite{tan2018mnasnet}.
However, NAS remains an intrinsically costly problem with 
a combinatorially large search space: \textit{e.g.}, searching for a 
ConvNet with 22 layers and five candidate operations per layer 
yields $5^{22} \approx 10^{15}$ possible networks.

\begin{figure}[t!]
  \centering
  \includegraphics[width=1.0\columnwidth]{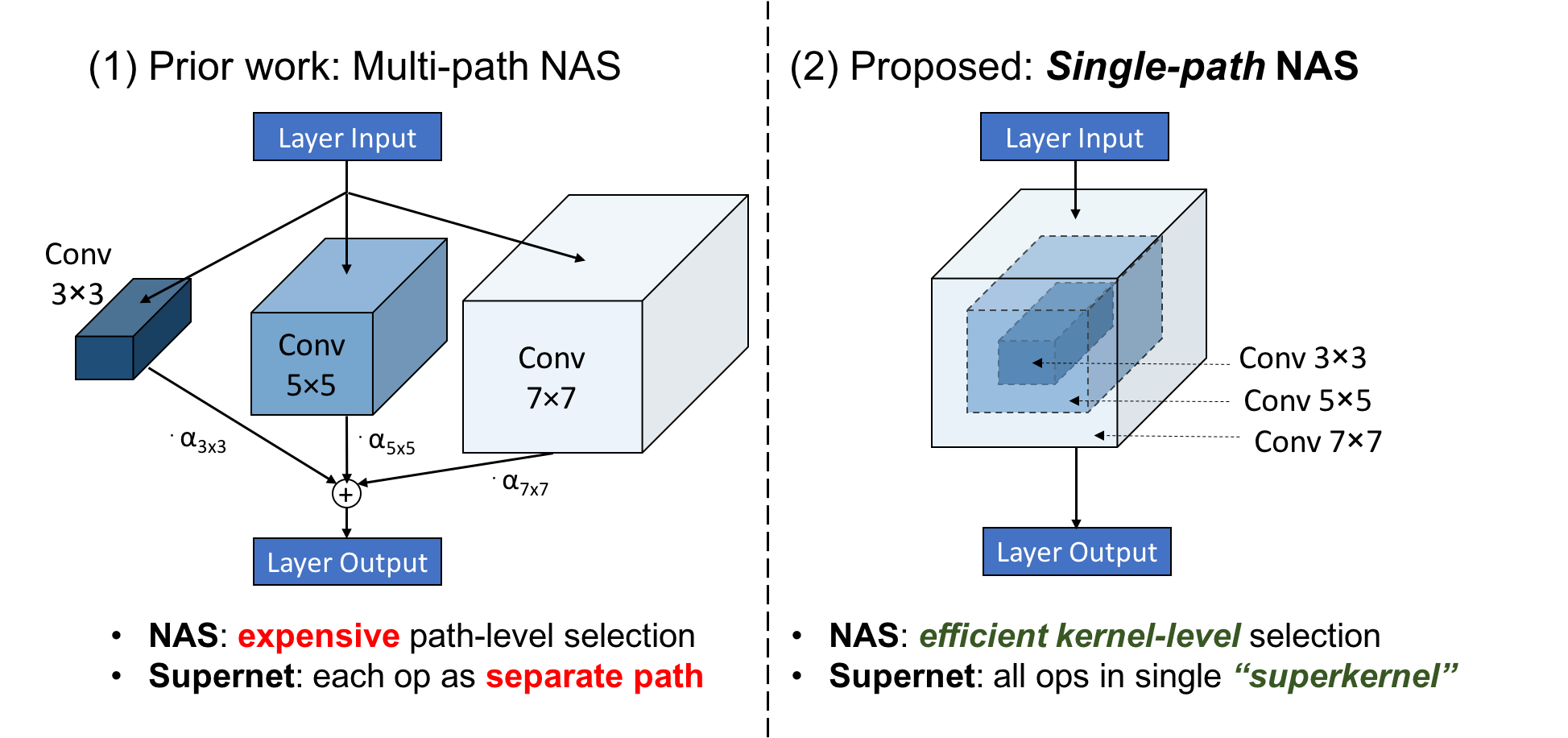}
  \vspace{-20pt}
  \caption{\textit{Single-Path NAS} directly 
  optimizes for the subset of convolution weights of 
  an over-parameterized ``\textbf{superkernel}'' in each ConvNet layer
  (right). Our \textbf{novel view} of the design space
  eliminates the need for maintaining separate paths for each candidate 
  operation, as in previous \textit{multi-path} approaches (left).}
  \vspace{-10pt}
  \label{fig:key_idea}
\end{figure}

\textbf{Inefficiencies of \textit{multi-path} NAS}:
Recent methods use one-shot formulations~\cite{liu2018darts,pham2018efficient}
by viewing the NAS problem as an operation/path 
selection problem: first, an over-parameterized, \textit{multi-path} 
supernet is constructed, where, for each layer, every candidate operation is 
added as a \textit{separate} trainable path (Figure~\ref{fig:key_idea}, left).
Next, NAS searches for the paths of the \textit{multi-path} supernet that 
yield the optimal architecture.
As expected, naively branching out all paths is inefficient, since 
the number of trainable parameters
during the search grows linearly with respect to the number 
of candidate operations per layer~\cite{bender2018understanding}.
To tame the memory explosion due to the \textit{multi-path} supernet,
current methods employ ``workaround'' solutions:
\textit{e.g.}, searching on a proxy dataset~\cite{wu2018fbnet}, 
or employing a memory-wise scheme where only few paths are 
updated during search~\cite{cai2018proxylessnas}. Nevertheless, these 
methods remain considerably costly, with total computational demand of
at least 200 GPU-hours. 

\begin{figure*}[ht!]
  \centering
  \includegraphics[width=1.0\textwidth]{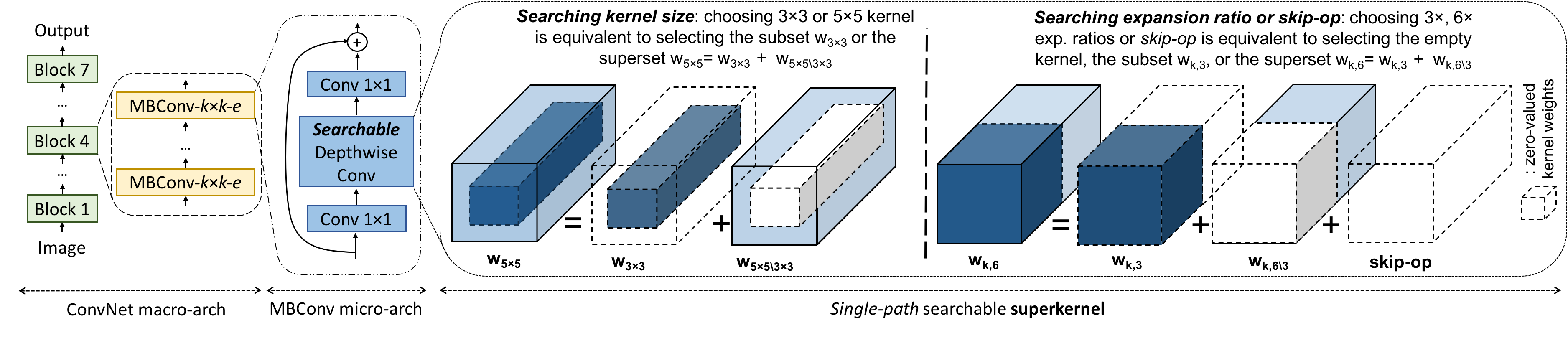}
  \vspace{-25pt}
  \caption{\textit{Single-path} NAS builds upon
  \textit{hierarchical} MobileNetV2-like search 
  spaces~\cite{tan2018mnasnet} to identify the mobile inverted bottleneck 
  convolution (MBConv) per layer (left).
  Our \textit{one-shot supernet} encapsulates all possible NAS architectures 
  in the search space, \textit{i.e.},
  different kernel size (middle) and expansion ratio (right) values, 
  without the need for appending each candidate operation
  as a separate path. \textit{Single-Path} NAS directly searches over the weights of 
  the per-layer \textbf{searchable ``superkernel''} that encodes all MBConv types.}
  \label{fig:design_space}
\end{figure*}

In this paper, we propose \textit{Single-Path NAS}, a novel NAS method for designing
hardware-efficient ConvNets in \textbf{less than 4 hours}. Our \textbf{key insight}
is illustrated in Figure~\ref{fig:key_idea} (right). 
Our key observation is that different candidate 
convolutional operations can be viewed as subsets of the weights of 
an over-parameterized \textbf{single ``superkernel''}. 
Instead of choosing from different paths/operations as 
in \textit{multi-path} methods, we view the NAS problem as 
\textit{finding which subset of the per-layer \textbf{``superkernel''} 
weights to use}, hence searching across 
a \textbf{single-path} one-shot NAS supernet.

\section{Proposed Method: \textit{Single-Path} NAS}

\textbf{Search space}: 
We build upon the MobileNetV2-based macro-architecture (Figure~\ref{fig:design_space}, 
left), where layers are grouped into blocks
based on their filter sizes~\cite{sandler2018mobilenetv2}. Each layer is a mobile 
inverted bottleneck convolution (conv) 
MBConv-$k\times k$-$e$~\cite{sandler2018mobilenetv2}, (\textit{i.e.}, 
a point-wise $1\times 1$ conv with expansion ratio $e$, 
a $k\times k$ depthwise conv, and a linear $1\times 1$ conv; Figure~\ref{fig:design_space}, left).
Our goal is to identify the MBConv-$k\times k$-$e$ type per layer.

\begin{table*}[t!]
\caption{\textit{Single-Path} NAS achieves state-of-the-art image classification
accuracy (\%) on ImageNet 
for similar on-device latency setting compared to previous 
NAS methods ($\leq 80 ms$ on Pixel 1), with up to 
$5,000 \times$ reduced search cost in terms of number of epochs. *The search cost
in epochs is estimated based on the claim~\cite{cai2018proxylessnas}
that ProxylessNAS is $200 \times$ faster than MnasNet. $\ddag$ChamNet does not detail
the model derived under runtime constraints~\cite{dai2018chamnet} 
so we cannot retrain or measure the latency.}
\begin{center}
\begin{small}
\begin{sc}
\scalebox{0.95}{
\begin{tabular}{l|cccc}
\toprule
Method & Top-1 Acc (\%) & Top-5 Acc (\%) & Runtime (ms) & Search  Cost (epochs) \\
\midrule
  
MobileNetV2~\cite{sandler2018mobilenetv2} & 72.00 & 91.00 & 75.00 & \multirow{2}{*}{-}  \\
MobileNetV2 (our impl.) & 73.59 & 91.41 & 73.57 &  \\\hline

Random search     & 73.78 $\pm$ 0.85 & 91.42 $\pm$ 0.56 & 77.31 $\pm$ 0.9 ms & - \\\hline

MnasNet~\cite{tan2018mnasnet} & 74.00 & 91.80 & 76.00 & \multirow{2}{*}{40,000}  \\
MnasNet (our impl.)  & 74.61 & 91.95 & 74.65 &   \\\hline
ChamNet-B~\cite{dai2018chamnet}   & 73.80 & -- & -- & 240$\ddag$  \\\hline
ProxylessNAS-R~\cite{cai2018proxylessnas} & 74.60 & 92.20 & 78.00 & \multirow{2}{*}{200*}  \\
ProxylessNAS-R (our impl.)  & 74.65 & 92.18 & 77.48 &   \\\hline
FBNet-B~\cite{wu2018fbnet} & 74.1 & - & - & \multirow{2}{*}{90}  \\
FBNet-B (our impl.)  & 73.70 & 91.51 & 78.33 &   \\\hline
\hline
\textit{Single-Path} NAS (\textbf{proposed}) & \textbf{74.96} & \textbf{92.21} & 79.48 & \textbf{8} (\textbf{3.75 hours})  \\
\bottomrule
\end{tabular}
}
\end{sc}
\end{small}
\end{center}
\vskip -0.1in
\label{tab:imagenet-sota}
\end{table*}

\textbf{``Superkernel''-based formulation}: Our key insight is that the 
candidate operations can be viewed as subsets of the \textbf{``superkernel''} weights. 
Without loss of generality, we denote the weights of two candidate operations,
\textit{e.g.}, $3 \times 3$ conv or $5 \times 5$ conv, as 
$\textbf{w}_{3 \times 3}$ and $\textbf{w}_{5 \times 5}$, respectively. 
We observe that $\textbf{w}_{3 \times 3}$ can be viewed as 
the \textit{inner} core of the $\textbf{w}_{5 \times 5}$ kernel, 
while ``zeroing'' out the ``\textit{outer}'' shell
$ \textbf{w}_{5 \times 5 \setminus 3 \times 3}$ (Figure~\ref{fig:design_space}, middle).
Thus, we can write the NAS decision as:
\begin{equation}
    \label{eq:proposed-form-2}
    \textbf{w}_{k} = \textbf{w}_{3 \times 3} + \mathbbm{1}(\left\Vert 
    \textbf{w}_{5 \times 5 \setminus 3 \times 3} \right\Vert^2 > t_{k=5}) 
    \cdot \textbf{w}_{5 \times 5 \setminus 3 \times 3}
\end{equation}
where $\mathbbm{1}(\cdot)$ is the indicator function that encodes the 
architectural NAS choice and $t_{k=5}$ is a latent variable 
(\textit{e.g.}, a threshold value) that controls the decision.

\textbf{Trainable NAS decisions}: Drawing inspiration from 
quantization decisions~\cite{ding2019flightnns}, we use the \textit{group Lasso term}
in the $\mathbbm{1}(\cdot)$ condition. Instead of picking the thresholds 
(\textit{e.g.}, $t_{k=5}$) by hand, we seamlessly treat them as trainable parameters to 
learn via gradient descent. To compute the gradients for thresholds, we relax the indicator 
function $g(x,t) = \mathbbm{1}(x>t)$ to a 
sigmoid function $\sigma(\cdot)$, when computing gradients, \textit{i.e.}, 
$\hat{g}(x,t) = \sigma(x>t)$. 

\textbf{Searching for expansion ratio or skip-op}: Since the 
kernel-based $\textbf{w}_{k}$ result (Equation~\ref{eq:proposed-form-2}) is a 
kernel itself, we can in turn apply our formulation to encode 
expansion ratio decisions, where $e=3$ or $e=6$ correspond to using
half or all the channels of an MBConv-$k\times k$-$6$ layer, respectively
(Figure~\ref{fig:design_space}, right). Finally, by ``zeroing'' out all channels, 
we encode the NAS decision of dropping the entire layer: 
\begin{equation}
    \begin{split}
    \label{eq:proposed-form-3}
    \textbf{w} = \mathbbm{1}(\left\Vert 
    \textbf{w}_{k, 3} \right\Vert^2 > & t_{e=3}) \cdot 
    ( \textbf{w}_{k, 3} + \\ & \mathbbm{1}(\left\Vert 
    \textbf{w}_{k, 6 \setminus 3} \right\Vert^2 > t_{e=6}) 
    \cdot \textbf{w}_{k, 6 \setminus 3} )
    \end{split}
\end{equation}
Hence, our \textbf{searchable superkernel} can sufficiently capture 
any MBConv type in the MobileNetV2-based design space (Figure~\ref{fig:design_space}).
For  input $\textbf{x}$, the output of the $i$-th MBConv layer of the network is
$o^i (\textbf{x}) = \text{conv}(\textbf{x}, \textbf{w}^{i} | t_{k=5}^{i}, t_{e=6}^{i}, t_{e=3}^{i})$.

\textbf{Differentiable NAS}: 
To account for both the accuracy and inference latency of the searched ConvNet,
we use a latency-aware formulation for the NAS problem~\cite{wu2018fbnet}:
\begin{equation}
    \label{eq:loss}
    \underset{\textbf{w}}{\text{min }}
    CE(\textbf{w} | \textbf{t}_{k}, \textbf{t}_{e} ) + \lambda \cdot 
    \text{log}(R(\textbf{w} | \textbf{t}_{k}, \textbf{t}_{e}))
\end{equation}
where $CE$ is the cross-entropy loss of the single-path model and $R$ is the 
runtime in milliseconds ($ms$) of the searched NAS model on the target device. The 
coefficient $\lambda$ modulates the trade-off between cross-entropy and runtime.

\textbf{Runtime model}: Prior art has showed that the on-device ConvNet runtime
can be modeled as the sum of each $i$-th layer's 
runtime $R(\textbf{w} | \textbf{t}_{k}, \textbf{t}_{e}) = 
\sum_{i} R^{i}(\textbf{w}^{i} | \textbf{t}^{i}_{k}, 
\textbf{t}^{i}_{e})$~\cite{wu2018fbnet,cai2017neuralpower,stamoulis2018hyperpower}.
To preserve the differentiability of the 
objective, we formulate the per-layer $R^{i}$ as a function of the 
NAS decisions. We profile the target mobile device (Pixel 1
smartphone) and we record the runtime for each candidate kernel operation per layer $i$. 
As a function of the expansion ratio decisions, we write:
\begin{equation}
    \label{eq:runtime-layer-e}
    \begin{split}
    R^{i}_{e} & = \mathbbm{1}(\left\Vert 
    \textbf{w}_{k, 3} \right\Vert^2 > t_{e=3}) \cdot 
    ( R^{i}_{5 \times 5, 3} + \\ & \mathbbm{1}(\left\Vert 
    \textbf{w}_{k, 6 \setminus 3} \right\Vert^2 > t_{e=6}) 
    \cdot ( R^{i}_{5 \times 5, 6} - R^{i}_{5 \times 5, 3} ))
    \end{split}
\end{equation}
By incorporating the kernel size decision, the runtime is:
\begin{equation}
    \label{eq:runtime-layer}
    \begin{split}
    R^{i} = &  \frac{R^{i}_{3 \times 3,6}}{R^{i}_{5 \times 5,6}} \cdot R^{i}_{e} + \\
    R^{i}_{e} & \cdot (1-\frac{R^{i}_{3 \times 3,6}}{R^{i}_{5 \times 5,6}}) \cdot 
    \mathbbm{1}(\left\Vert 
    \textbf{w}_{5 \times 5 \setminus 3 \times 3} \right\Vert^2 > t_{k=5})
    \end{split}
\end{equation}
As in Equation~\ref{eq:proposed-form-2}, we relax the indicator 
function to a sigmoid function $\sigma(\cdot)$ when computing gradients.
To evaluate the prediction accuracy of the runtime model, 
we generate 100 random ConvNets and we measure their runtime
on the device. Our model can accurately predict the actual runtimes:
the Root Mean Squared Error (RMSE) is $1.32ms$, which corresponds to an 
average $1.76\%$ prediction error. 

\section{Experiments}
\label{sec:results}

\textbf{Experimental Setup}: We select Pixel 1 as the target 
device since it allows for a representative 
comparison with prior work that optimizes for this platform. 
We run our framework using \texttt{TensorFlow (TF)} on 
TPUs-v2~\cite{jouppi2017datacenter}. We deploy the ConvNets 
on the device with \texttt{TF TFLite}.
We profile on-device runtime using the Facebook AI Performance Evaluation 
Platform~\cite{faipep}.
We implement our trainable ``superkernels'' on \texttt{Keras}. 

We apply our method to design ConvNets for image classification 
on ImageNet~\cite{deng2009imagenet} running on Pixel 1
with an overall target latency of $80ms$. We train the derived 
\textit{Single-Path} NAS model for 350 epochs. We summarize 
the results in Table~\ref{tab:imagenet-sota}. To enable a representative 
comparison of the search cost per method, we directly report 
the number of epochs per method, hence canceling out the 
effect of different hardware systems (GPU vs TPU hours).

\textbf{State-of-the-art on-device image classification}:
\textit{Single-Path} NAS achieves top-1 accuracy of $\textbf{74.96\%}$, which is 
the new state-of-the-art ImageNet accuracy among hardware-efficient NAS methods. 
More specifically, \textbf{our method achieves better top-1 accuracy than ProxylessNAS
by} $+\textbf{0.31}\%$, while maintaining on par target latency of $\leq 80ms$ on the 
same target mobile phone. \textit{Single-Path} NAS outperforms methods in 
this mobile latency range, \textit{i.e.}, better than MnasNet ($+0.35\%$), 
FBNet-B ($+0.86\%$), and MobileNetV2 ($+1.37\%$).

\textbf{Reduced search cost}: \textit{Single-Path} NAS has a total search cost
of 8 epochs, which is $\textbf{5,000} \times$ faster than MnasNet, $\textbf{25} \times$ 
faster than ProxylessNAS, and $\textbf{11} \times$ faster than FBNet.
Specifically, MnasNet reports a total of 40k train epochs. In turn, ChamNet
trains an accuracy predictor on 240 samples. ProxylessNAS reports $200 \times$ 
search cost improvement over MnasNet, hence the overall cost is the TPU-equivalent of 
200 epochs. Finally, FBNet reports 90 epochs. Overall, we 
search for \textbf{$\sim 10k$ steps} (8 epochs with a batch size of $1024$),
which corresponds to total wall-clock time of \textbf{3.75 hours} on a TPUv2.
In particular, given than a TPUv2 has 2 chips with 4 cores each, this 
corresponds to a total of 30 TPU-hours.

\section{Discussion \& Future Work}

\textbf{Novel idea}: The key insight behind our work is to revisit the one-shot 
\textbf{supernet} NAS design space with a \textit{single-path} view, by 
formulating the NAS problem as \textit{finding which subset of kernel weights
to use} in each ConvNet layer. While concurrent works consider relaxed convolution
formulations~\cite{shin2018differentiable,hundt2019sharpdarts,guo2019single}, 
they either use design spaces and objectives that have been shown to be 
hardware inefficient (\textit{e.g.}, cell-based space, FLOP count), or 
they do not intrinsically relax the kernel over both kernel-size and
channels dimensions. 

\textbf{Future work}: The efficiency of our \textit{single-path}
design space could enable future work beyond our differentiable
NAS formulation and based on reinforcement learning or evolutionary 
methods. Moreover, our methodology can be flexibly extended 
to other hardware design goals, \textit{e.g.}, power, memory, energy,
and communication constraints~\cite{dong2018dpp,stamoulis2018designing,stamoulis2018designing}.
To this end and to foster reproducibility, unlike all recent 
hardware-efficient NAS methods which release pretrained 
models only, we open-source and fully document our method 
at: \url{https://github.com/dstamoulis/single-path-nas}.

\bibliography{singlepathnas}
\bibliographystyle{icml2019}

\end{document}